%% file: nmi-version.tex
\documentclass[12pt]{article}
\usepackage{arxiv}
\usepackage[utf8]{inputenc} 
\usepackage[T1]{fontenc}    
\usepackage{hyperref}       
\usepackage{url}            
\usepackage{booktabs}       
\usepackage{amsfonts}       
\usepackage{amsmath}
\usepackage{nicefrac}       
\usepackage{microtype}      
\usepackage{cleveref}       
\usepackage{lipsum}         
\usepackage{graphicx}
\usepackage{lineno}
\usepackage{natbib}
\usepackage{pdflscape}
\usepackage{rotating}
\usepackage[table]{xcolor}
\usepackage{doi}
\usepackage{multirow}
\usepackage{multicol}
\usepackage{subcaption}
\usepackage{tabularx}
\usepackage{rotating}
\usepackage{comment}
\usepackage{float}
\title{Rethinking Evaluation in the Era of Time Series Foundation Models: (Un)known Information Leakage Challenges}

\newif\ifuniqueAffiliation

\ifuniqueAffiliation 
\author{ \href{https://orcid.org/0009-0005-9136-8525}{\includegraphics[scale=0.06]{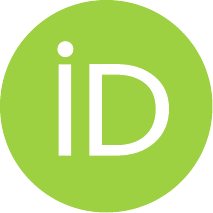}
\hspace{1mm}Marcel Meyer}\\
	\And
	\href{https://orcid.org/0000-0002-8411-6347}{\includegraphics[scale=0.06]{orcid.pdf}\hspace{1mm}Sascha Kaltenpoth} \\
    \And
	\href{https://orcid.org/0009-0000-8515-5711}{\includegraphics[scale=0.06]{orcid.pdf}\hspace{1mm}Kevin Zalipski} \\
    \And
	\href{https://orcid.org/0000-0002-0369-1607}{\includegraphics[scale=0.06]{orcid.pdf}\hspace{1mm}Oliver Müller} \\
}
\else
\usepackage{authblk}

\newbox{\orcid}\sbox{\orcid}{\includegraphics[scale=0.06]{orcid.pdf}} 
\author[1]{%
	\href{https://orcid.org/0009-0005-9136-8525}{\usebox{\orcid}\hspace{1mm}Marcel Meyer\thanks{\texttt{marcel.meyer@uni-paderborn.de}}}%
}
\author[1]{%
	\href{https://orcid.org/0000-0002-8411-6347}{\usebox{\orcid}\hspace{1mm}Sascha Kaltenpoth}%
}
\author[1]{%
	\href{https://orcid.org/0009-0000-8515-5711}{\usebox{\orcid}\hspace{1mm}Kevin Zalipski}%
}
\author[1]{%
	\href{https://orcid.org/0000-0002-0369-1607}{\usebox{\orcid}\hspace{1mm}Oliver Müller}%
}
\affil[1]{Paderborn University, Data Analytics Group}
\fi

\renewcommand{\headeright}{}
\renewcommand{\undertitle}{}
\renewcommand{\shorttitle}{\textit{arXiv} Template}

\hypersetup{
pdftitle={Rethinking Evaluation in the Era of Time Series Foundation Models: (Un)known Information Leakage Challenges},
pdfsubject={TSFM Benchmarking, Time Series Foundation Models},
pdfkeywords={First keyword, Second keyword, More},
}

\date{}
\begin{document}

\maketitle

\begin{abstract}
Time Series Foundation Models (TSFMs) represent a new paradigm for time-series forecasting, promising zero-shot predictions without the need for task-specific training or fine-tuning. However, similar to Large Language Models (LLMs), the evaluation of TSFMs is challenging: as training corpora grow increasingly large, it becomes difficult to ensure the integrity of the test sets used for benchmarking. An investigation of existing TSFM evaluation studies identifies two kinds of information leakage: (1) train-test sample overlaps arising from the multi-purpose reuse of datasets and (2) temporal overlap of correlated train and test series. Ignoring these forms of information leakage when benchmarking TSFMs risks producing overly optimistic performance estimates that fail to generalize to real-world settings. We therefore argue for the development of novel evaluation methodologies that avoid pitfalls already observed in both LLM and classical time-series benchmarking, and we call on the research community to adopt principled approaches to safeguard the integrity of TSFM evaluation.
\end{abstract}

\section{Introduction}
\label{introduction}

Time Series Foundation Models (TSFMs) represent an emerging paradigm in forecasting, drawing inspiration from the architecture and training methodologies of foundation models in natural language processing (NLP). In contrast to traditional time series models, TSFMs are pre-trained on large time series corpora, enabling zero-shot forecasting without task-specific adaptation \citep{liangFoundationModelsTime2024}. Over the last years, a highly dynamic landscape with a rapidly growing family of TSFMs has eveolved \citep{ansariChronosLearningLanguage2024,auerTiRexZeroShotForecasting2025,cohenThisTimeDifferent2025,dasDecoderonlyFoundationModel2024}.

Yet, the very property that makes these TSFMs powerful, training globally on the world's time series, creates structural evaluation problems, akin to issues recently observed in the evaluation of large language models (LLMs). In the NLP domain, training on vast portions of the internet has given rise to an ``evaluation crisis" \citep{liaoRethinkingModelEvaluation2023}, in which test set contamination \citep{mirzadehGSMSymbolicUnderstandingLimitations2024, ravautHowMuchAre2024, liOpenSourceDataContamination2024} and memorization effects \citep{changSurveyEvaluationLarge2024} have led to overly optimistic performance estimates. As this perspective will show, current TSFM evaluations are vulnerable to analogous issues, leading to important implications for fair and reliable forecasting evaluation in the era of TSFMs.

While novel benchmark strategies such as benchmarking on held-out data \citep{aksuGIFTEvalBenchmarkGeneral2024} and clean train-test splits \citep{qiuTFBComprehensiveFair2024} have been proposed recently, these approaches cannot address the fundamental sources of information leakage in TSFMs. We identify two such sources (see Section \ref{sec:challenges}): First, \textbf{direct information leakage}, arising from the multi-purpose reuse of public datasets across model training and evaluation pipelines. Second, \textbf{indirect information leakage}, arising from temporal overlap between correlated training and test series that often share a common causal driver — such as the COVID-19 pandemic simultaneously distorting correlated financial time series across geographies. Together, these sources risk turning current TSFM benchmarks into measures of memorization rather than generalization.

Our investigation reveals that information leakage is already present in TSFM benchmarking. Tracing the dataset lineage of 22 published TSFMs reveals no community consensus on which data should be used for training versus evaluation: in several documented cases, one model's pre-training corpus is another model's test set, making direct cross-model comparison nearly impossible. Yet, even benchmarks that carefully avoid such direct overlap remain exposed to indirect information leakage: a model trained on global stock indices through 2020 can learn the COVID-19 crash and exploit that structure when asked to forecast any correlated series from the same period, even one it has never directly seen (see Section \ref{sec:challenges}).

The TSFM field risks repeating, in compressed time, the evaluation crisis that has undermined trust in LLM benchmarking. Preventing this doesn't require incremental fixes to existing benchmarks, but a principled rethinking of what valid evaluation means when a model has been trained on the world's time series.

\section{Time Series Foundation Models: The Current State of Evaluation}
Understanding the current state of TSFM evaluation requires first appreciating what distinguishes these models architecturally from classical forecasting methods, and how the benchmarking practices developed for those methods have been adapted to assess them.

\subsection{Time Series Foundation Models: Global Training, Zero-Shot Inference}
\label{sec:tsfm-definition}
The foundation model paradigm, originally defined as training on broad data at scale to enable adaptation across downstream tasks \citep{bommasaniOpportunitiesRisksFoundation2021}, found its most prominent expression in LLMs such as the GPT and Llama families \citep{openaiGPT4TechnicalReport2023,grattafioriLlama3Herd2024}, whose performance scales predictably with training data, parameters, and compute \citep{kaplanScalingLawsNeural2020}. Recognizing the structural sequence similarity between language modeling and time series analysis, recent work has adapted this paradigm to forecasting, producing a growing family of Transformer-based architectures that range from encoder-decoder designs \citep{ansariChronosLearningLanguage2024} to decoder-only \citep{dasDecoderonlyFoundationModel2024} and encoder-only models \citep{goswamiMOMENTFamilyOpen2024}. Since the emergence of TSFMs a few years ago, 22 models have been presented at top venues including TMLR, ICML, NeurIPS, ICLR, and WWW, as shown in Table~\ref{tab:unified_models}.

\begin{table}[h]
\centering
\caption{Time series foundation models published in recent years}
\label{tab:unified_models}
\begin{tabular}{llll}
\toprule
\textbf{Model Name} & \textbf{Author} & \textbf{Year} & \textbf{Conference/Journal} \\
\midrule
ForecastPFN        & \citet{dooleyForecastPFNSyntheticallyTrainedZeroShot2023}     & 2023 & NeurIPS 23 \\
GPT4TS             & \citet{zhouOneFitsAll2023}               & 2023 & NeurIPS 23 \\
Lag-Llama          & \citet{rasulLagLlamaFoundationModels2024}        & 2024 & NeurIPS 23 Workshop R0-FoMo \\
LLMTime            & \citet{gruverLargeLanguageModels2024}           & 2024 & NeurIPS 23 \\
Time-LLM           & \citet{jinTimeLLMTimeSeries2024}           & 2024 & ICLR 24 \\
Chronos            & \citet{ansariChronosLearningLanguage2024}         & 2024 & TMLR \\
TimesFM            & \citet{dasDecoderonlyFoundationModel2024}       & 2024 & ICML 24 \\
Moirai             & \citet{wooUnifiedTrainingUniversal2024}            & 2024 & ICML 24 \\
MOMENT             & \citet{goswamiMOMENTFamilyOpen2024}         & 2024 & ICML 24 \\
Timer              & \citet{liuTimerGenerativePretrained2024}              & 2024 & ICML 24 \\
UniTime            & \citet{liuUniTimeLanguageEmpoweredUnified2024}            & 2024 & WWW 24 \\
TinyTimeMixer      & \citet{ekambaramTinyTimeMixers2024}         & 2024 & NeurIPS 24 \\
Time-MoE           & \citet{shiTimeMoEBillionScaleTime2024}           & 2025 & ICLR 25 \\
Moirai-MoE         & \citet{liuMoiraiMoEEmpoweringTime2024}         & 2024 & ICML 25 \\
VisionTS           & \citet{chenVisionTSVisualMasked2024}          & 2024 & ICML 25 \\
LightGTS           & \citet{wangLightGTSLightweightGeneral2025}          & 2025 & ICML 25 \\
Sundial            & \citet{liuSundialFamilyHighly2025}            & 2025 & ICML 25 \\
TimesFM-IC         & \citet{fawInContextFineTuningTimeSeries2025}           & 2025 & ICML 25 \\
ROSE               & \citet{wangGeneralTimeSeries2025}           & 2025 & ICML 25 \\
FlowState          & \citet{grafFlowstateSamplingRate2025}         & 2025 & NeurIPS 25 Workshop BERT2S \\
TiRex              & \citet{auerTiRexZeroShotForecasting2025}             & 2025 & NeurIPS 25 \\
Toto               & \citet{cohenThisTimeDifferent2025}             & 2025 & NeurIPS 25 \\
\bottomrule
\end{tabular}
\end{table}

Beyond the Transformer backbone, TSFMs draw on a range of architectural approaches: LLM-based models encode time series values directly as language tokens \citep{gruverLargeLanguageModels2024,xuePromptCastNewPromptBased2024}, vision-based models represent them as grayscaled images \citep{chenVisionTSVisualMasked2024}, and reprogramming-based architectures mix textual and continuous representations \citep{jinTimeLLMTimeSeries2024}. Despite this diversity, a key commonality runs through almost all TSFMs: pre-training on large, heterogeneous collections of time series that appear to follow scaling laws \citep{edwardsScalinglawsLargeTimeseries2024}, enabling zero-shot forecasting across a wide range of domains, frequencies, and horizons without retraining (MOMENT is the notable exception, requiring task-adaptive fine-tuning of its output layer \citep{goswamiMOMENTFamilyOpen2024}). This zero-shot capability differs fundamentally from its NLP counterpart: while LLMs generalize to entirely different tasks \citep{brownLanguageModelsAre2020}, TSFMs generalize across domains, series, input lengths, and horizons, extending to new time series rather than new problem types \citep{ansariChronosLearningLanguage2024, dasDecoderonlyFoundationModel2024, wooUnifiedTrainingUniversal2024}.

\subsection{Classical and TSFM Benchmarking Practices}
\label{sec:benchmarking-methodologies}
Traditional statistical or machine-learning-based models for time series forecasting generally require explicit training on the target time series. These local models exhibit only limited generalization capabilities; that is, they can only predict values that are identically distributed to their training data. Consequently, the dominant logic in time series benchmarking relies on \textit{time-based} train/test splits \citep{hyndmanForecastingPrinciplesPractice2021}. This approach creates a temporal separation between training and evaluation data to prevent information leakage and is employed in major time series collections such as the Monash repository \citep{godahewaMonashTimeSeries2021} and TFB \citep{qiuTFBComprehensiveFair2024}.

\begin{figure}[h!]
    \centering
    \includegraphics[width=1\linewidth]{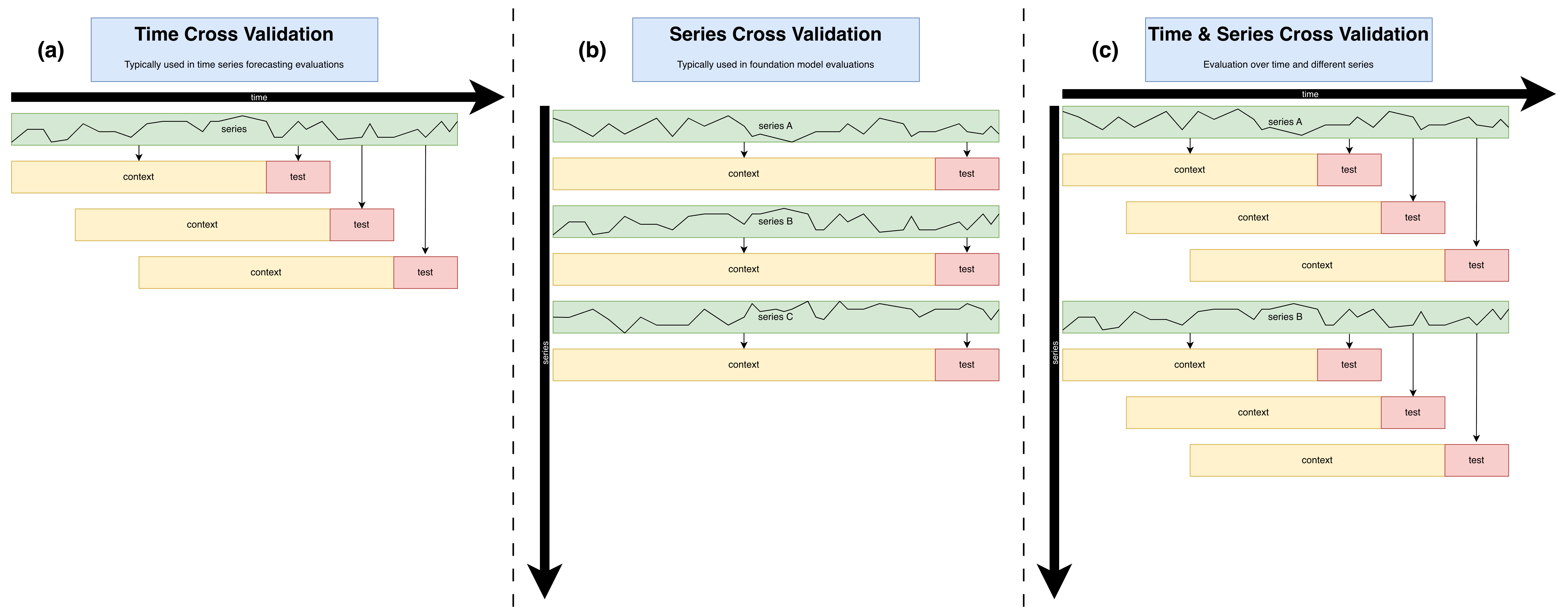}
    \caption{Evaluation strategies along the time (left) or the domain dimension (right).}
    \label{fig:eval_current}
\end{figure}

To rigorously assess a model's predictive capabilities under evolving temporal dynamics, \textit{Time-Series Cross-Validation (TSCV)} is widely regarded as best practice. Instead of using a single time-based split, TSCV uses the idea of rolling windows (see Figure \ref{fig:eval_current} (a)). This evaluation strategy does not necessarily require retraining; instead, the training set can be fixed while the inference window is rolled forward, reducing evaluation costs to inference-only computation \citep{hewamalageForecastEvaluationData2023}.

Forecasting competitions, like the M-competitions or Kaggle, try to avoid the risk of test set contamination by soliciting forecasts for unpublished \textit{private test sets} \citep{bojerKaggleForecastingCompetitions2021,makridakisM5CompetitionBackground2022}. These competitions are typically held at irregular intervals. The recent M6 competition took this further, introducing live evaluation using future values of financial assets as the target, bypassing reliance on historical data entirely \citep{makridakisM6ForecastingCompetition2024}.

While current TSFM benchmarking studies also rely on historical data for evaluation, their logic diverges from the above described time-series cross-validation strategy. Instead of using TSCV on one series as illustrated in Figure \ref{fig:eval_current} (a), they typically evaluate models using many series but with only a single time-based split per series \citep{dasDecoderonlyFoundationModel2024, gruverLargeLanguageModels2024}. Figure \ref{fig:eval_current} (b) shows the focus of most TSFM evaluations. Here, the focus is not on temporal depth, but on generalization across heterogeneous series, especially in zero-shot scenarios where complete datasets or domains are held out of the pre-training \citep{shiTimeMoEBillionScaleTime2024, ansariChronosLearningLanguage2024}. This approach is focused on the generalization capabilities of TSFMs and prevents cherry-picking of test sets. Two prominent examples of this strategy are TSFM-Bench, which comprises 21 test sets with time series from diverse domains (e.g., energy, finance), statistical characteristics (e.g., stationary, trend), and frequencies (e.g., quarter-hourly, hourly) \cite{liTSFMBenchComprehensiveUnified2025} and GIFT-Eval, which comprises 23 curated test sets with time series of seven domains, ten frequencies, and multivariate inputs \citep{aksuGIFTEvalBenchmarkGeneral2024}.

While this series cross validation focuses on evaluating the generalization capabilities of TSFMs, it is not without drawbacks. \citet{roqueCherryPickingTimeSeries2025} showed that the final ranking of models depends heavily on which series are selected as test sets. With just four cherry-picked test sets, out of the distribution of all available test sets, 46\% of the benchmarked models could appear as state-of-the-art purely due to selection bias. Notably, deep learning models were found to be more susceptible to this variance than classical baselines. Especially when evaluating general-purpose TSFMs, generalization is a core evaluation focus.

Figure \ref{fig:eval_current} (c) illustrates the current best practice in TSFM evaluation, that is, a combined time and series cross validation \citep{yuTemporalRegularizedMatrix2016, salinasDeepARProbabilisticForecasting2020}. While this approach has first adopters, notably \citet{goktas2025tempusbench,shchurFevbenchRealisticBenchmark2025}, it is not yet commonly used in TSFM benchmarking.

\subsection{Information Leakage in TSFM Benchmarking: A Recognized Challenge}
As pre-training corpora grow larger and more heterogeneous, ensuring that evaluation data has not been seen during training becomes increasingly difficult. The same dynamic has been observed in LLM evaluation: models pre-trained on large internet crawls such as Common Crawl\footnote{https://commoncrawl.org/} or The Pile\footnote{https://pile.eleuther.ai/} had already seen many benchmark tasks during pre-training \citep{carliniQuantifyingMemorizationNeural2022,changSurveyEvaluationLarge2024}, ultimately contributing to what has been described as an LLM ``evaluation crisis'' \citep{liaoRethinkingModelEvaluation2023}. TSFMs face the same structural problems, and the community has begun to respond. As shown in the GIFT-Eval ablation studies, information leakage can strongly inflate evaluation performance \citep{aksuGIFTEvalBenchmarkGeneral2024}, and several recent benchmarks have developed explicit strategies to limit its impact \citep{aksuGIFTEvalBenchmarkGeneral2024, goktas2025tempusbench, xuFidelTSHighFidelityMultimodal2025}. GIFT-Eval pairs an explicit pre-training corpus of approximately 230 billion data points with a dedicated test set of 177 million data points \citep{aksuGIFTEvalBenchmarkGeneral2024}. TempusBench draws on 48 forecasting tasks from sources currently absent from any known pre-training corpus, and further proposes live evaluation on real-world data \citep{goktas2025tempusbench}. Fidel-TS takes a similar approach through restricted live APIs \citep{xuFidelTSHighFidelityMultimodal2025}.

Yet, none of these efforts have been preceded by a systematic analysis of where information leakage actually originates from, nor of how many distinct forms it takes. Without that foundation, benchmark design remains reactive rather than principled.

\section{Two Distinct Pathways of Information Leakage in TSFM Evaluation}
\label{sec:challenges}

The integrity of any benchmark rests on a strict separation between training and evaluation data. For TSFMs, this separation is threatened along two distinct dimensions. The first is direct: datasets used for pre-training or fine-tuning reappear as test sets, undermining evaluation. The second is indirect and specific to time series: temporally overlapping but nominally disjoint series can be statistically correlated, allowing information about the test period to leak into the training signal without any test data point ever being seen directly. Evidence for both forms of leakage can already be found in current TSFM evaluations.

\subsection{When Pre-Training Flexibility Comes at the Cost of Benchmark Comparability}
\label{sec:used-data-in-tsfm}
Each TSFM team independently selects which datasets to use for pre-training, fine-tuning, and evaluation. Rightly so, since the freedom to train on diverse, large-scale corpora is central to the foundation model paradigm. Yet this very flexibility carries a structural side effect that has received little attention: once a dataset has been used for pre-training or fine-tuning by any model, it is no longer admissible as a foundation for benchmarking models against others. With each new model trained on a new combination of datasets, the set of valid datasets for benchmarking is reduced. A lineage analysis of 22 TSFMs published at leading machine learning venues (TMLR, ICML, NeurIPS, ICLR, WWW) and their workshops through January 2026 (Table \ref{tab:unified_models}) makes this concrete. For each model, every dataset was categorized into one of three roles: \textit{pre-training} (used only for initial model training and thus excluded from evaluation), \textit{train/test} (split temporally for fine-tuning and in-domain evaluation), or \textit{zero-shot} (reserved exclusively for held-out evaluation on unseen data).

\begin{figure*}[ht]
    \centering
    \includegraphics[width=1.00\textwidth]{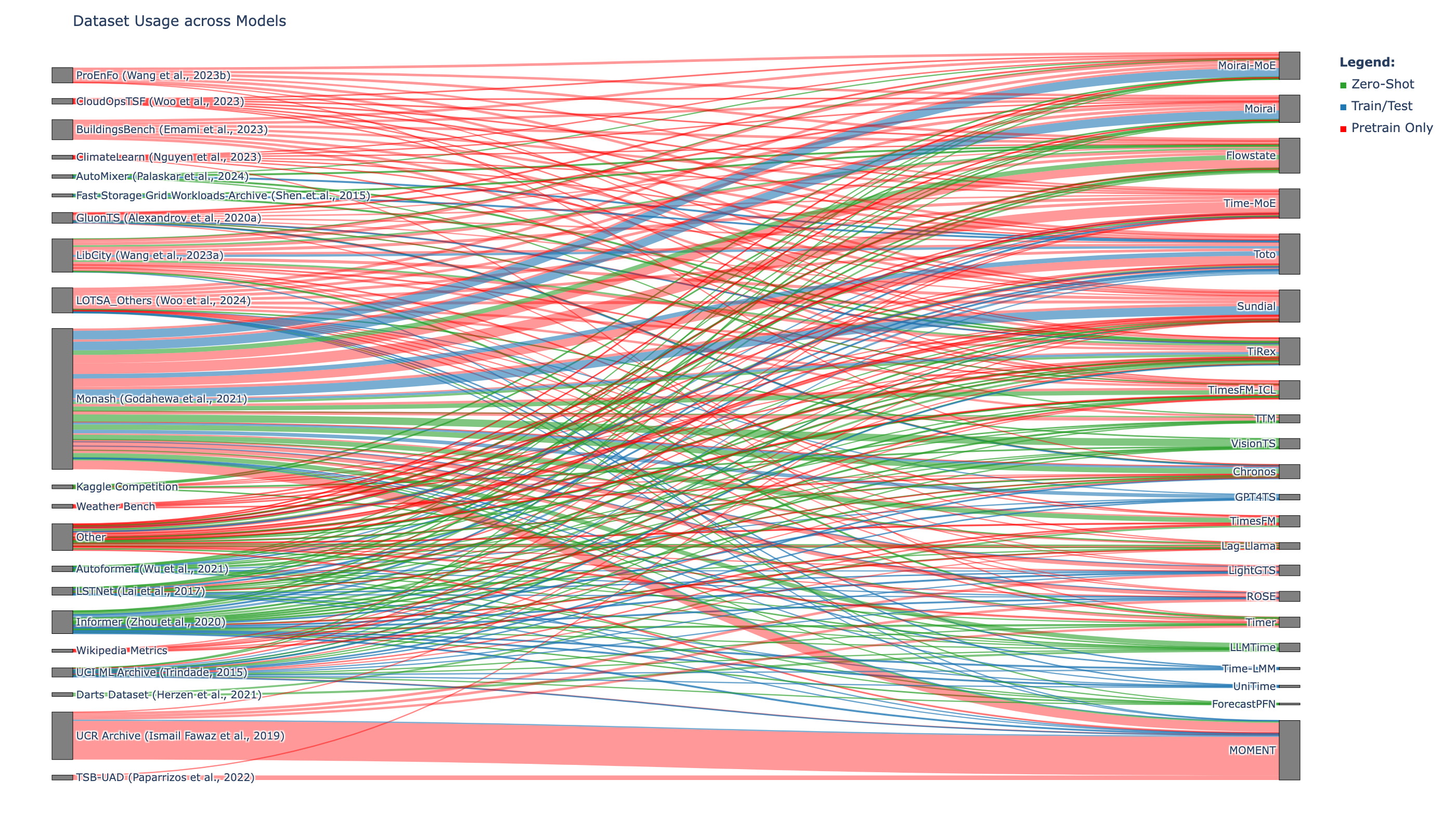}
    \caption{Lineage of dataset collections (left) used for training and evaluating recent Time Series Foundation Models (right). Typically, a collection contains multiple different datasets. Lines indicate cases where at least one dataset of a collection was used for pre-training, train/test, or zero-shot evaluation. Smaller datasets (used n<10) are shown as "Other". }
    \label{fig:datasets-train-eval}
\end{figure*}

The lineage map (Figure \ref{fig:datasets-train-eval}; full dataset-level analysis in Section \ref{sec:data-availability}) confirms this: across 401 datasets identified, each TSFM team assembled a distinct combination of pre-training, train/test, and zero-shot datasets, and no single dataset has been universally reserved for evaluation. Only 6\% of these 401 datasets have never appeared in any model's pre-training or fine-tuning corpus. These are the only candidates that could, in principle, support genuine zero-shot comparisons across the full set of published TSFMs.

The pervasiveness of this problem is best illustrated through concrete examples. The \textit{Australian Electricity Demand} dataset from the Monash collection \citep{godahewaMonashTimeSeries2021} has been used for pre-training (e.g., Lag-Llama, Timer), train/test evaluation (e.g., Moirai), and zero-shot forecasting (e.g., Chronos, TimesFM), making it practically useless for any cross-model comparison. Similarly, the \textit{Informer} collection \citep{zhouInformerEfficientTransformer2021}, widely treated as a zero-shot benchmark, contains individual series such as ETTh1, ETTh2, and ETTm1 that have also been used for pre-training (e.g., in Lag-Llama) or train/test evaluation (e.g., in UniTime).

These overlaps are often difficult to detect, because datasets are routinely remixed, renamed, and redistributed across repositories. For instance, the dataset \textit{``Elecdemand''} from the Monash Repository is a scaled subset ($1/1000$) of \textit{``Australian Electricity Demand''}. The dataset \textit{``ElectricityLoadDiagrams20112014''} appears under the name \textit{``Electricity''} in both the Autoformer and Monash collections \citep{godahewaMonashTimeSeries2021, wuAutoformerDecompositionTransformers2022} and as \textit{``ECL''} in the Informer collection \citep{zhouInformerEfficientTransformer2021}. Even datasets sharing the same name may contain different underlying data: the \textit{car ride shares} dataset is included from the same year 2015 in both the Monash and GluonTS collections. Although both repositories cite the same GitHub source\footnote{\url{https://github.com/fivethirtyeight/uber-tlc-foil-response}}, a close examination reveals that they contain different time series of distinct For-Hire Vehicles (Uber in GluonTS vs.\ Lyft in Monash). Consequently, it is necessary to always compare the underlying data itself, keeping in mind that it could have been transformed in various ways (scaling, handling missing values, renaming, resampling, etc.).

A particularly telling example of how dataset provenance can become convoluted through multiple transformations and repositories involves the ``Solar'' dataset. The Chronos pre-training corpus includes solar power data from all US states at two frequencies: 5T (5 minutes) and 1H (hourly), sourced directly from the original data provider\footnote{\url{https://www.nlr.gov/grid/solar-power-data}}. GIFT-Eval, however, includes only the Alabama subset as a test set, aggregated to 10T, 1H, 1D (daily), and 1W (weekly), sourced through the LSTNet dataset repository\footnote{\url{https://github.com/laiguokun/multivariate-time-series-data/tree/master/solar-energy}}. LSTNet also sourced the data from the original provider but, as noted, included only Alabama rather than all states. In addition, the Alabama solar subset appears in the Monash Repository\footnote{\url{https://forecastingdata.org/}} at frequencies 10T and 1W. This modified version was used as pretraining data by TinyTimeMixers and Lag-Llama \citep{ekambaramTinyTimeMixers2024, rasulLagLlamaFoundationModels2024}. While TinyTimeMixers and Lag-Llama were published before GIFT-Eval, the authors of Sundial validate against GIFT-Eval and state they excluded any overlaps without explicitly listing which datasets they excluded \citep{aksuGIFTEvalBenchmarkGeneral2024, liuSundialFamilyHighly2025}. The authors of TiRex and Toto discovered this entanglement and excluded the dataset as a precaution, though TiRex explicitly notes the difficulty of verifying whether the data across repositories is actually identical \citep{auerTiRexZeroShotForecasting2025, cohenThisTimeDifferent2025}.

The consequences of undetected contamination can be severe. As summarized in Supplementary Information Section \ref{app:leakage-investigations}, documented cases of leakage between training and test sets have led to over 50\% better test scores. Several such cases suggest that overlaps between train and test data are not always identified during the peer-review process \citep{saravananAnalyzingPerformanceTime2024, montetBenchmarkingFoundationModels2025, liFoundTSComprehensiveUnified2024}. This is not surprising: exactly determining on which datasets a model has been trained requires careful reading of the original paper, its appendix, and sometimes even analyzing published source code. During the lineage analysis, a potential overlap surfaced between the pre-training and test set in TimesFM \citep{dasDecoderonlyFoundationModel2024}, as the \textit{traffic hourly} dataset appears to have been used during pretraining while also being included in the reported Monash evaluation benchmark. This anecdotal example underscores how difficult it is to maintain a reliable overview of datasets; a challenge that only intensifies as the number of published TSFMs grows.

\subsection{When Temporal Overlap Turns Correlation Into Leakage}
\label{sec:correlation_leakage}
Time series carry a special structural property: temporal correlation between nominally independent series. This correlation often has causal roots, yet TSFMs do not learn causal relationships; they learn from observed statistical patterns alone. When an exogenous shock affects multiple domains simultaneously --- a pandemic, a financial crisis, a geopolitical shift --- it imprints a recognizable signature across many series at once. The same effect operates at a local scale: weather conditions at a specific location directly shape solar energy production \citep{aryandoustEnhancedSpatiotemporalElectric2022}, and local traffic patterns simultaneously influence air quality measurements and delivery time predictions. A model trained on any of these series during the affected period has, in effect, learned something about all of them, even those it never encountered directly. While spurious correlation without any shared cause could in principle produce a similar effect, it may also lead to worse predictions, as the model transfers patterns that bear no true relationship to the test series. But the systematic correlation arising from genuine common drivers is the more prevalent and serious concern.

\begin{figure}[h]
    \centering
    \includegraphics[width=0.75\linewidth]{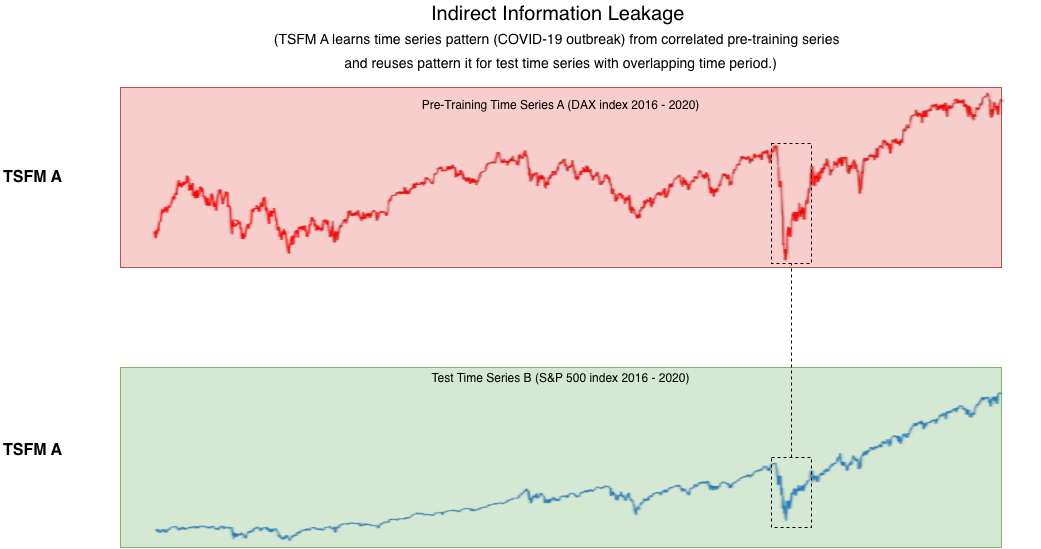}
    \caption{Indirect Information Leakage - Temporal Overlap of Correlated Train Series A and Test Series B.}
    \label{fig:global-patterns}
\end{figure}

The evaluation implication follows directly. A key strength of TSFMs lies in their ability to learn subtle patterns in a data-driven manner and transfer them to previously unseen series. At large scale, however, this same capability becomes a liability for evaluation. When correlated training and test series overlap temporally, a sufficiently flexible TSFM can transfer learned statistical structure from the former to the latter. Crucially, this does not reflect genuine forecasting skill: the model is not learning the generative dynamics of the test series, but exploiting correlation produced by a shared external driver. As illustrated in Figure~\ref{fig:global-patterns}, a TSFM trained on the DAX index through the COVID-19 crash absorbs the temporal signature of that event. When subsequently asked to forecast the S\&P 500 over the same window, it can reproduce that shape --- not by understanding US market dynamics, but because both indices were driven by the same global shock and are therefore highly correlated during that period. The key distinction from classical benchmarking is therefore not only \textit{which} series appeared in training, but \textit{when} those series were observed: contemporaneous observation of correlated series embeds information about the test period even when no test data point was ever directly seen. This temporal overlap violates the independence assumption fundamental to unbiased evaluation, inflating performance metrics in ways that do not reflect real-world forecasting ability.

This concern is not merely theoretical. \citet{rodrigoDataLeakagePretrained2024} provided a clean empirical demonstration using highly correlated public transport series from Madrid: metro, bus, road, and train ridership throughout 2024. The metro series was withheld entirely from training and served as the zero-shot forecasting target. For all remaining series, two training configurations were compared: in the first, a common train-test split date was applied, so that none of the correlated series extended into the metro test period; in the second, all non-metro series were kept in training up to the end of the evaluation window, overlapping temporally with the metro test period. Both models were then evaluated exclusively on the metro test period. The model whose correlated training series overlapped with the test period achieved a Mean Absolute Error (MAE) of 248.24 compared to 439.14 for the properly split model, a performance improvement of approximately 43\% attributable solely to this indirect form of leakage. No metro data point was ever seen during training; the advantage came entirely from the shared temporal dynamics of the correlated series.

The Madrid experiment used a near-ideal setting for leakage: virtually all training series were highly correlated with the test target. The more pressing question for TSFM evaluation is whether temporal leakage remains detectable when the leaked signal is diluted among a large, heterogeneous training corpus. A controlled replication around the COVID-19 stock market crash of early 2020 suggests the answer is yes. To isolate the effect cleanly, a Time-Series Transformer (TST) \citep{zerveasTransformerbasedFrameworkMultivariate2020} was trained from scratch, rather than relying on an existing TSFM that could already carry this form of leakage. The architecture shares multi-variate characteristics with recent models such as MOMENT or Chronos-2 \citep{goswamiMOMENTFamilyOpen2024,ansariChronos2UnivariateUniversal2025} while remaining small enough for fast training saturation. Two almost identical training sets were constructed\footnote{Details on the general setup, used datasets, scaling, hyperparameter-tuning and results can be found in Supplementary Information Section \ref{app:global_pattern_leakage_experiment}.}: both contained the same non-financial series (NN5 Daily and Tourism Monthly from the Monash collection) plus seven major stock indices (Bovespa, CAC40, DAX, DowJones, FTSE 100, Nikkei 225, TSX)\footnote{All stock index data was retrieved from the website \textit{investing.com}}, differing only in the temporal range of the stock data. The first set cut the stock series at December 2019, before the pandemic; the second extended them through December 2020, encompassing the crash and partial recovery. In both cases, the S\&P 500 was withheld entirely as the zero-shot test target. Stock prices were chosen deliberately: they are notoriously hard to model beyond short horizons \citep{goyalComprehensive2022Look2024}, ensuring that any systematic performance gap reflects leaked external information rather than improved modeling of intrinsic dynamics.

\begin{figure}[h!]
    \centering
    \includegraphics[width=\linewidth]{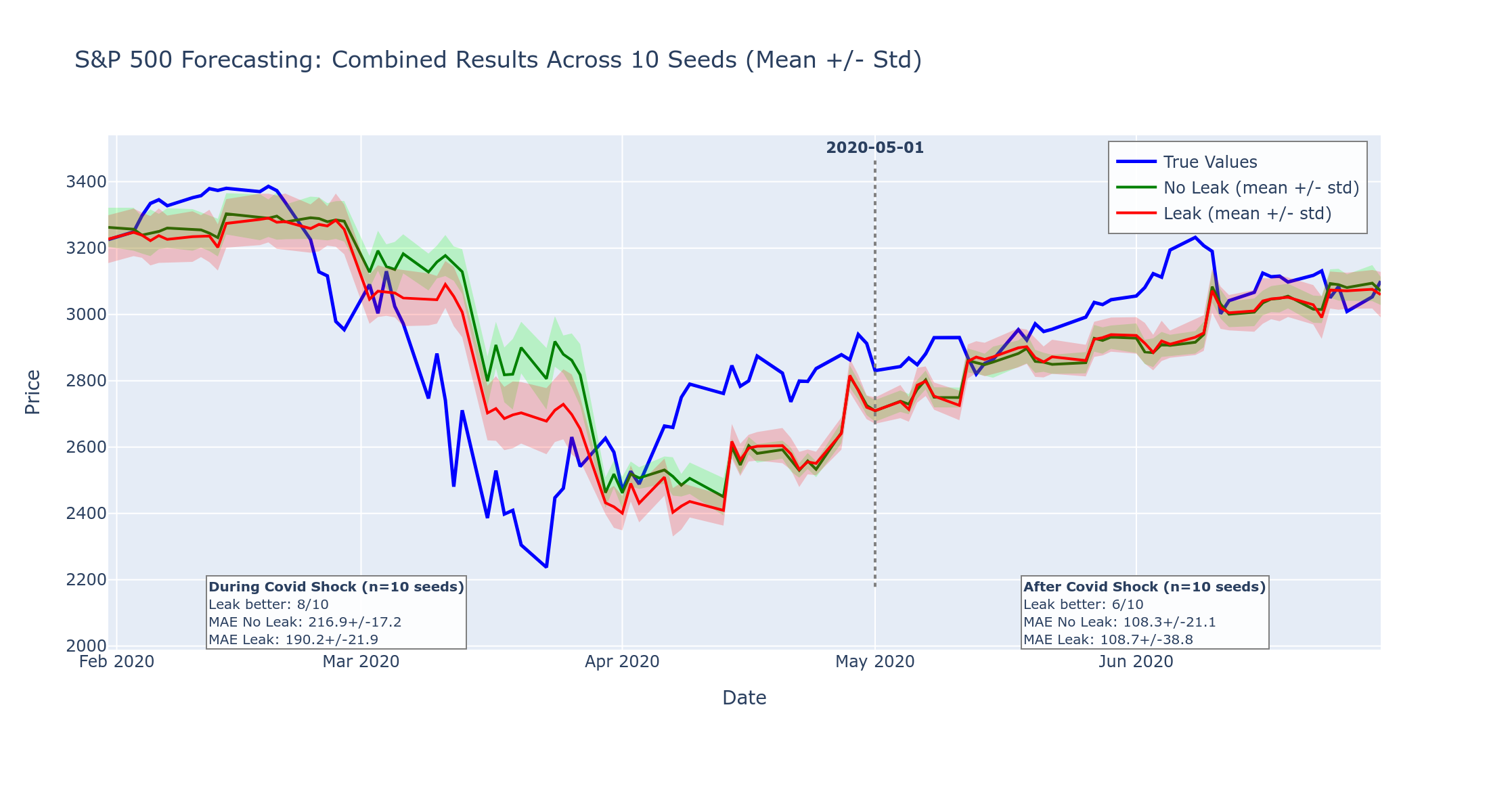}
    \caption{Zero-shot forecasting for S\&P 500 stock index during the Covid shock in the year 2020. In blue is the ground truth, in green a pretrained model with data up to December 2019, in red a model pretrained until December 2020. In both cases S\&P 500 is not part of the training data.}
    \label{fig:exp_seeds_mean_std}
\end{figure}

Figure \ref{fig:exp_seeds_mean_std} displays the mean and standard deviation of the predictions across ten random seeds for both models, with per-seed results provided in Supplementary Information Table \ref{tab:exp_single_seeds}. The results are unambiguous: during the crash period, the model trained on data with temporal overlap predicted the initial market decline significantly better than the clean model, outperforming it in eight out of ten runs with individual improvements reaching up to approximately 37\%. Up to the first of May, the temporal overlap model achieved an average MAE of about 190 compared to about 217 for the non-leakage model. For the remaining part, both models converged to very similar average performance (108.3 vs.\ 108.7), confirming that the advantage was specific to the period of shared causal influence rather than reflecting a general training benefit. This performance gap is all the more striking given that the leaked signal comprised just seven time series out of 484, or roughly 1,600 out of over 205,000 total training observations (Table \ref{tab:leakage_summary}). If a small transformer trained from scratch on fewer than 500 series can exploit temporal correlation for a measurable advantage, production-scale TSFMs trained on millions of series are likely more susceptible, not less, given their greater capacity to memorize subtle patterns and the higher probability that their pre-training corpora contain series that are temporally overlapped and correlated with any given test set.

Both the Madrid transport experiment \citep{rodrigoDataLeakagePretrained2024} and the COVID stock market experiment involve indirect leakage between series of the same frequency. Yet a further, largely undiscussed complication concerns the relationship between different frequency representations of the same underlying process. The community has not yet reached consensus on this matter: some treat different sampled frequencies as practically independent \citep{auerTiRexZeroShotForecasting2025,shchurFevbenchRealisticBenchmark2025, dasDecoderonlyFoundationModel2024}, while others leverage similar patterns across frequencies architecturally and consider them dependent \citep{liuMoiraiMoEEmpoweringTime2024, grafFlowstateSamplingRate2025}. We argue that frequency variants of the same underlying process should be treated as dependent, for two reasons. First, time series sampled at different frequencies from the same source retain very high correlation when mapped to common timestamps. Second, if a TSFM is capable of capturing the ``bigger picture'' of a time series (and this is precisely the promise of foundation models), it should recognize the same trends and seasonal patterns regardless of temporal resolution. Training on one frequency and testing on another from the same ground truth should therefore be considered a form of indirect leakage.

\section{Two Requirements for Information-Leakage-Free TSFM Evaluation}
\label{sec:benchmark-requirements}
The essence of time series forecasting is to predict the \textit{future}. Paradoxically, using data from the \textit{past} to evaluate their ability to do so is standard practice (Section \ref{sec:benchmarking-methodologies}). Before the advent of TSFMs, this paradox was not a serious issue, as traditional time series models have to be trained on past values of the same time series they are expected to continue. The only pitfall to avoid is adhering to the correct ordering of training and test observations.

\begin{figure}[h]
    \centering
    \includegraphics[width=1.0\linewidth]{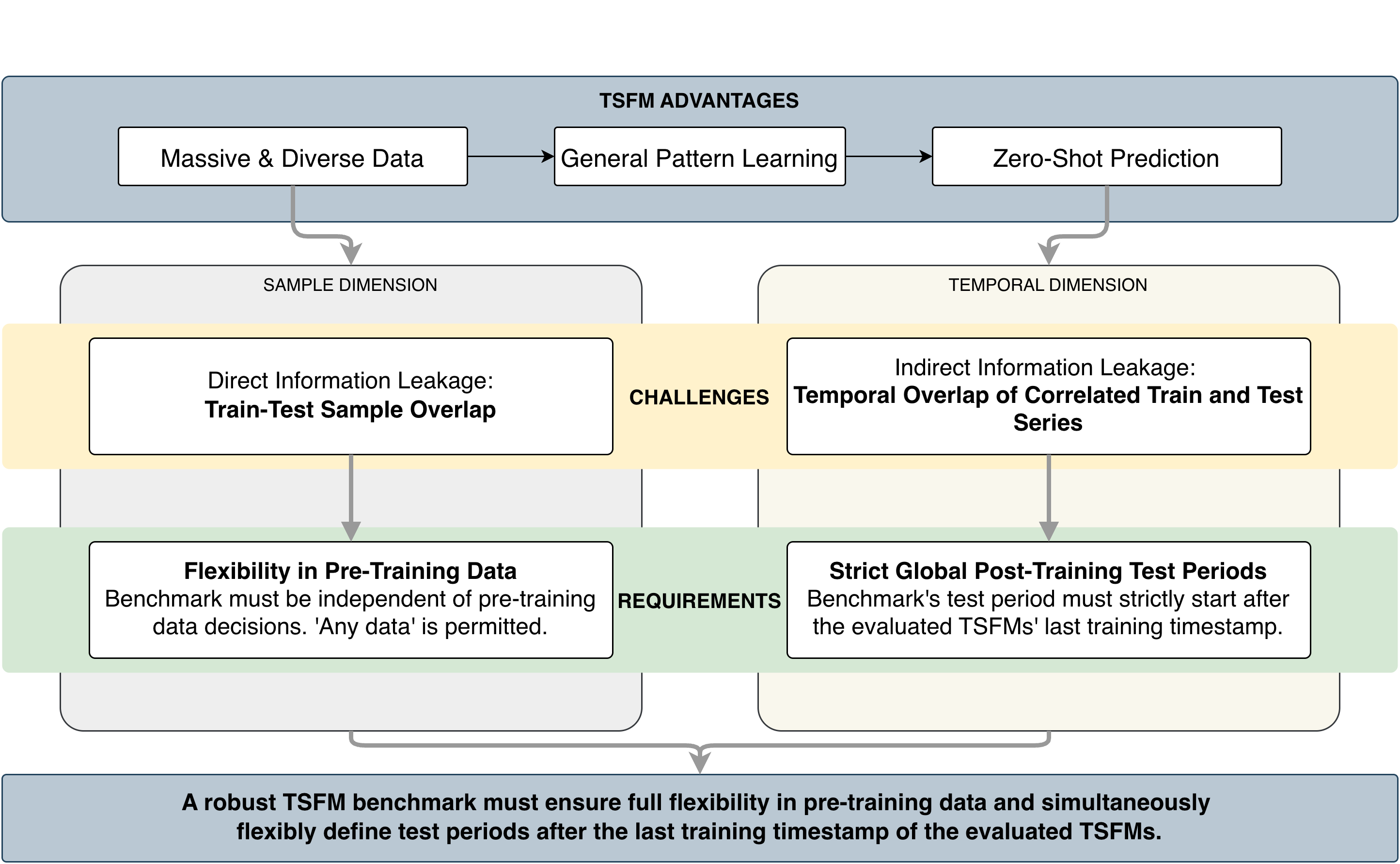}
    \caption{Information leakage risks and challenges arising from the advent of TSFMs, and the requirements needed to address them.}
    \label{fig:information_leakage_requirements}
\end{figure}

However, the paradigm shift behind TSFMs shatters this assumption. Unlike local models, TSFMs learn patterns globally. Consequently, strict temporal ordering within a single time series is no longer sufficient to guarantee a fair evaluation. To rigorously assess these models, we must address leakage across two distinct dimensions: through what was in the training data (the sample dimension) and through when the training data was observed (the temporal dimension) (see Figure \ref{fig:information_leakage_requirements}).

\textbf{Requirement 1: Full flexibility in pre-training data.} Attempting to prevent contamination by restricting what data TSFMs may train on is both counterproductive and unenforceable. Restrictions artificially limit the potential of foundation models, and the lineage analysis presented here demonstrates that verifying the absence of train-test overlap across opaque, large-scale pre-training corpora is practically impossible. The responsibility for preventing leakage must therefore shift from model developers to benchmark designers. A robust evaluation framework should be agnostic to pre-training choices, allowing any available data to be used for training. This means that test sets must be \textit{genuinely novel}: data that could not plausibly have appeared in any pre-training corpus. The benchmark, not the model, must guarantee separation.

\textbf{Requirement 2: Strict global post-training test periods.} Indirect temporal leakage cannot be prevented by controlling which series appear in training and test sets; it arises from correlation between any temporally overlapping series. A hard temporal barrier is therefore needed: if the latest pre-training timestamp across all evaluated models is $t$, then every observation in the test set must come from $t+1$ or later. This barrier must be determined globally, based on the actual training cutoff of the latest model in the comparison, rather than imposed as a fixed date. Fixing a universal cutoff would require all researchers to align their training protocols to the same date and would prevent models from learning current patterns, both of which are undesirable. A flexible, model-aware temporal barrier $t_{now}$ avoids these drawbacks while ensuring that no test observation $t_{now}+1$ could have been influenced by any training observation through shared causal drivers.

Together, these two requirements form a coherent basement for new frameworks. Requirement~1 addresses the sample dimension of leakage by making test data provably novel. Requirement~2 addresses the temporal dimension by ensuring strict temporal precedence of all training over all test data. Neither requirement alone is sufficient; both are necessary. Realizing both demands not incremental adjustments to existing evaluation practices, but a structural rethinking of how TSFM benchmarks are designed and maintained.

\section{Conclusion}
We call on the community to develop new benchmark methodologies grounded in both requirements proposed here: full flexibility in pre-training data and strict global post-training test periods. Possible directions include constantly newly sourced test data, synthetic test data, or fast-paced competitions on unreleased and unbiased test data.

While these two requirements define the conditions for trustworthy evaluation, several critical questions remain open. To date, the effect size of test-set contamination, when a test dataset is present in the pre-training corpus, has only been shown to be significant in isolated cases. Likewise, the magnitude of temporal leakage arising from correlated and overlapping time series remains largely unknown in real TSFM settings and is likely to vary across domains, types of global events, and other contextual factors. Carefully designed TSFM training regimes will be required to rigorously quantify these effects.

Ultimately, we believe that the requirements we have established lay the foundation for robust, fair, and information-leakage-free benchmarking. This is essential for preventing an evaluation crisis for TSFMs akin to what has been observed with LLMs.

\section{Data Availability}
\label{sec:data-availability}
The resulting table of the TSFM data-lineage analysis is supplied in the Github Repository under \href{https://github.com/DAG-UPB/rethinking-tsfm-evaluation/blob/main/lineage-analysis/Time-Series-Foundation-Models-Lineage.csv}{this link}. For the indirect information leakage experiment in Section \ref{sec:correlation_leakage}, the used data is available through the source code as well via Github: \url{https://github.com/DAG-UPB/rethinking-tsfm-evaluation}

\section{Code Availability}
The code to reproduce the indirect information leakage experiment in Section \ref{app:global_pattern_leakage_experiment} is provided via Github: \url{https://github.com/DAG-UPB/rethinking-tsfm-evaluation}

\bibliography{references}
\bibliographystyle{icml2025} 

\include{supplementary_information}

\end{document}

%% file: supplementary_information.tex
\appendix

\section{Leakage Investigations}
\label{app:leakage-investigations}
\subsection{Cases of Overlaps in Train-Test Samples}

This section summarizes individual cases where train-test sample overlaps unintentionally led to information leakage in recent benchmarking studies. Through our lineage analysis, we could easily compare the used benchmark datasets in the study against the TSFM training datasets (for full lineage table see Section \ref{sec:data-availability}).

In one case, the benchmark creators, due to the TSFMs' incomprehensible datasets, unintentionally included three evaluation datasets as test sets that had already been used for the pretraining of TimesFM, UniTS, and TTM \citep{liFoundTSComprehensiveUnified2024}. Based on our analysis of the benchmark results, this lead to an 47\% - 184\% lower mean squared error (MSE) rate compared to best models not pre-trained on the leaking datasets. The advantage of the best TSFM on non-leaked datasets is only between 0.3\% and 14\%. This performance benefit through information leakage appears to align with findings from the Moirai leakage example (see Section \ref{app:moirai-leakage}). In another example a peer-reviewed paper benchmarked TSFM models on the Electricity dataset, which has been used for pretraining of every of tested TSFMs \cite{saravananAnalyzingPerformanceTime2024}. Also in the energy domain, a paper benchmarked TSFMs on the Spanish dataset (among others), which is also included in the pretraining data of the tested TSFMs. Chronos achieved a sMAPE of 4.854 where the best non-TSFM model TiDE achieved a sMAPE of 8.102 \cite{montetBenchmarkingFoundationModels2025}. Taken together, these illustrative case highlight the increasing difficulties for finding suitable benchmarking data for TSFMs.

\subsection{Intended Information-Leakage Experiment for Moirai Model}
\label{app:moirai-leakage}

\begin{table*}[h!]
\label{tab:train/moirai-leakage}
\caption{Information leakage in TSFM comparing Moirai with (Moirai Leakage) and without (Moirai) information leakage during training phase for different model sizes. Metric is MAPE. Based on experiments by \citet{aksuGIFTEvalBenchmarkGeneral2024}.}
\begin{sc}
\centering
\resizebox{!}{0.10\textheight}{%
\begin{tabular}{lll|cc|cc|cc}
\toprule
 &  & \multicolumn{1}{r|}{Horizon} & \multicolumn{2}{c|}{Short} & \multicolumn{2}{c|}{Medium} & \multicolumn{2}{c}{Long} \\
 & & & \multirow{3}{*}{Moirai Leakage} & \multirow{3}{*}{Moirai}
            & \multirow{3}{*}{Moirai Leakage} & \multirow{3}{*}{Moirai}
            & \multirow{3}{*}{Moirai Leakage} & \multirow{3}{*}{Moirai} \\
Dataset & Frequency & Model Size \;\;\; &  &  &  &  &  &  \\
\midrule
 \multirow{3}{*}{loop\_seattle} & \multirow{3}{*}{5T} & S & \textbf{0.84} & 0.87 & \textbf{0.75} & 0.77 & \textbf{0.70} & 0.75 \\
  &  & B & \textbf{0.67} & 0.84 & \textbf{0.42} & 0.83 & \textbf{0.50} & 0.78 \\
  &  & L & \textbf{0.66} & 0.83 & \textbf{0.33} & 0.85 & \textbf{0.46} & 0.81 \\
\cline{1-9}
 \multirow{3}{*}{loop\_seattle} & \multirow{3}{*}{H} & S & 1.22 & \textbf{1.19} & 0.73 & \textbf{0.70} & \textbf{0.70} & 0.71 \\
  &  & B & \textbf{0.96} & 1.08 & \textbf{0.54} & 0.65 & \textbf{0.49} & 0.59 \\
  &  & L & \textbf{0.84} & 0.89 & \textbf{0.53} & 0.71 & \textbf{0.47} & 1.18 \\
\cline{1-9}
 \multirow{3}{*}{m\_dense} & \multirow{3}{*}{H} & S & 0.73 & \textbf{0.70} & \textbf{0.70} & 0.71 & \textbf{0.71} & 0.91 \\
  &  & B & \textbf{0.54} & 0.65 & \textbf{0.49} & 0.59 & \textbf{0.53} & 0.61 \\
  &  & L & \textbf{0.53} & 0.71 & \textbf{0.47} & 1.18 & \textbf{0.49} & 1.69 \\
\cline{1-9}
 \multirow{3}{*}{sz\_taxi} & \multirow{3}{*}{15T} & S & \textbf{0.95} & 1.11 & 0.65 & \textbf{0.60} & \textbf{2.12} & 2.30 \\
  &  & B & 0.90 & \textbf{0.84} & 0.71 & \textbf{0.64} & \textbf{2.16} & 2.42 \\
  &  & L & \textbf{0.78} & 0.82 & 0.71 & \textbf{0.60} & \textbf{2.14} & 2.24 \\
\cline{1-9}
\bottomrule
\end{tabular}
}

\end{sc}
\end{table*}

\cite{aksuGIFTEvalBenchmarkGeneral2024} performed an empirical analysis using the Moirai TSFM to demonstrate the effects of information leakage between training and test sets. The authors prepared a new pre-training and held-out evaluation dataset without any overlaps and trained the Moirai architecture on it. In addition, they took an already pretrained Moirai model from a previous publication, whose pre-training dataset contained 0.1\% of the newly defined held-out evaluation data, resulting in a deliberately small amount of information leakage. The leaked data includes nine datasets and three different forecast horizons (short, medium, and long), though not all combinations of datasets and horizons are present. 

\begin{figure}[ht]
    \centering
    \includegraphics[width=1\linewidth]{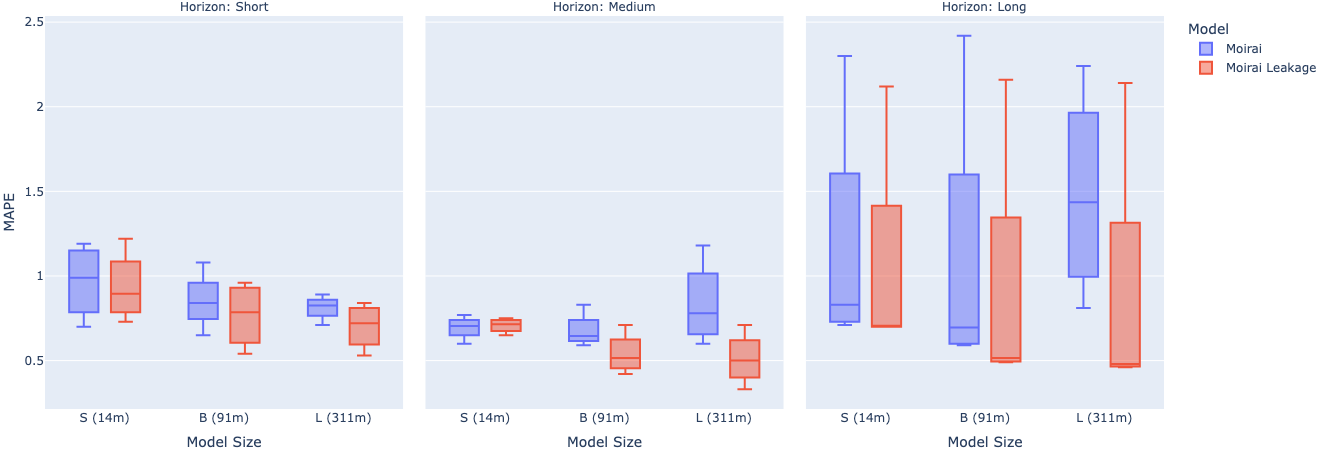}
    \caption{Comparing the model Moirai once with information leakage (Moirai Leakage), where the model was exposed to the test data during pre-training and the same model trained without information leakage (Moirai). Short, Medium and Long forecasting horizons were investigated. Model sizes are S (small), B (base) and L (large) and compared by the MAPE metric (the lower the better). Based on experiments by \citet{aksuGIFTEvalBenchmarkGeneral2024}.}
    \label{fig:moirai-leakage}
\end{figure}

Figure \ref{fig:moirai-leakage} summarizes the MAPE of both models. On short horizons, \textit{Moirai Leakage} has on average an 8 percentage points lower MAPE score across all model sizes. At medium horizons, \textit{Moirai Leakage} exhibits an average accuracy advantage of 15 percentage points. And on long horizons, \textit{Moirai Leakage} achieves an average MAPE score that is 29 percentage points lower than the leakage-free \textit{Moirai}.

Evaluation across model sizes reveals that larger models benefit more substantially from data leakage. The small model (S) shows a modest average improvement of 4 percentage points, while the base model (B) demonstrates a more pronounced advantage of 13 percentage points. Most notably, the large model (L) exhibits the strongest leakage effect with an average MAPE reduction of 34 percentage points across all forecasting horizons, indicating that model capacity amplifies the impact of training-test data contamination.

These results vividly demonstrate the significant impact that \textit{train-test sample overlap} can have on the performance of TSFMs and show that larger models are especially prone to this type of leakage.

\subsection{Potential Overlaps of Correlated Train and Test Series in Dataset Collections}
\begin{figure}[ht]
    \centering
    \includegraphics[width=\linewidth]{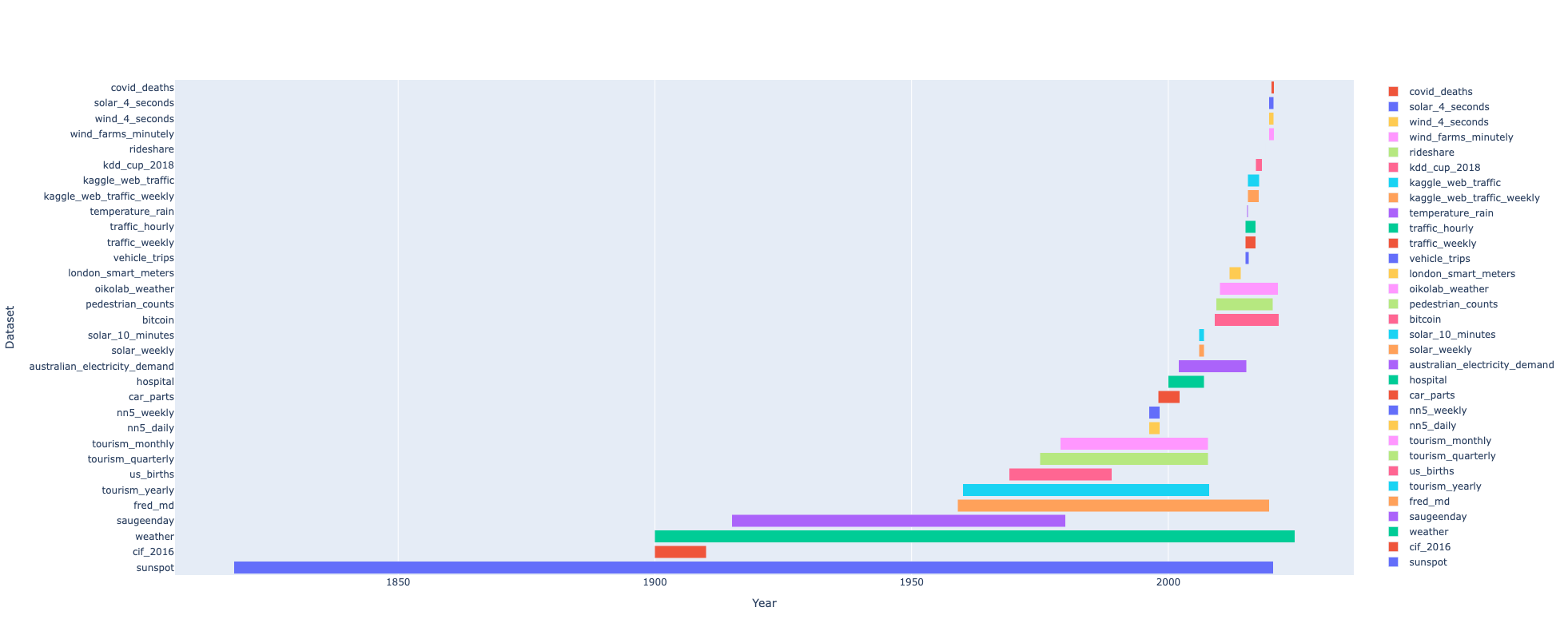}
    \caption{Date range of datasets within Monash dataset collection. Note that the ``cif\_2016" dataset was originally delivered without a timestamp and the start year was defined as 1900. No data cleaning was done.}
    \label{fig:monash_time}
\end{figure}

It may be theoretically possible that existing model evaluation results have already been influenced by overlaps of correlated train and test series. We analyse the overlapping time frames in the Monash Dataset \footnote{\url{https://huggingface.co/datasets/Monash-University/monash_tsf}}
as can be seen in Figure \ref{fig:monash_time}. Test datasets shouldn't lie in the same time period as other training datasets. As an example, the proposed Australian Electricity Demand test time frame (April 2015) lays in the same period as the training period of the "Weather (Australia)" training set (1900-2021). The Moirai and Moirai-MoE models use both datasets for train/test forecasts, meaning if there are any global patterns in the weather data which could influence the Australian Electricity Demand data, this would be a kind of information leakage, at least on theoretical basis.
Similarly, the Oikolab Weather dataset ranges from January 2010 to May 2021, which is weather data located in Melbourne. The test set of the dataset Pedestrian Count (also Melbourne) is located at the end of April 2020. This test period lays within the training data of the Oikolab Weather, which could be a greater potential global pattern information leakage than in previous example. Both datasets are included in the datasets of Moirai, Time-Moe, Sundial and Toto.

Moreover, LagLlama uses the \textit{Beijing Multi-Site Air Quality} dataset from 12 districts in Beijing from 2013 to 2017 as a pre-training dataset and simultaneously uses the \textit{Beijing PM2.5} air quality dataset from the Beijing airport between 2010 and 2013, which could also in theory lead to leakage.

\section{Temporal Overlap of Correlated Series Experiment}
\label{app:global_pattern_leakage_experiment}

\subsection{Data visualizations and statistics}
\begin{figure}[h!]
    \centering
    \includegraphics[width=1\linewidth]{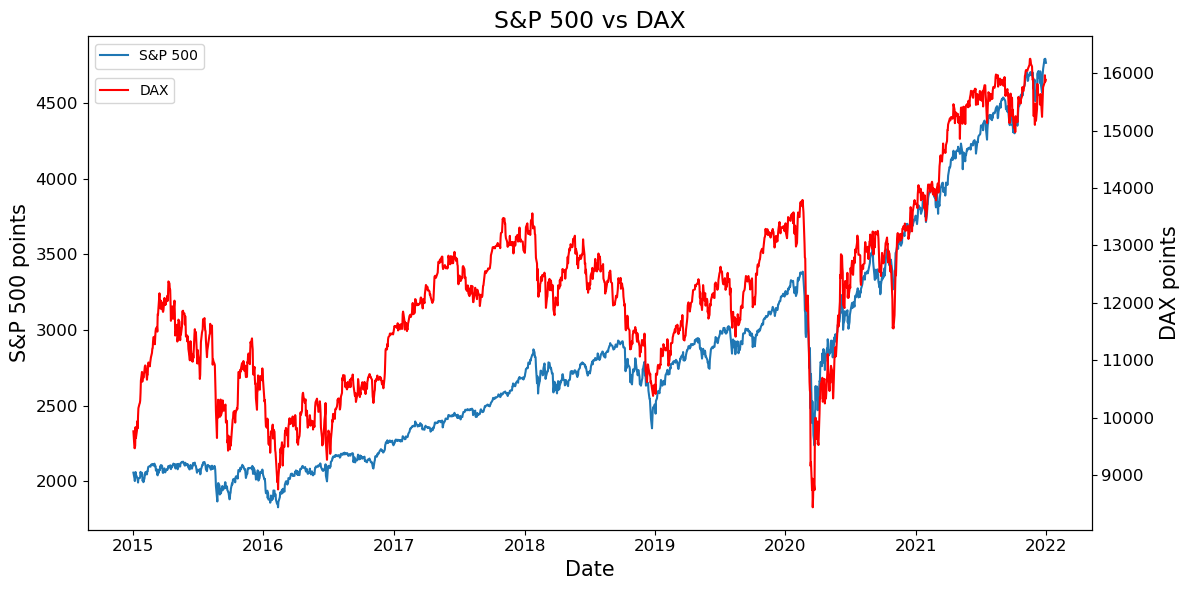}
    \caption{Comparison of DAX and S\&P 500 points over time}
    \label{fig:dax_sp_comp}
\end{figure}

Figure \ref{fig:dax_sp_comp} shows two example time series from major stock indices, the DAX and S\&P 500. Despite representing different markets, both series display remarkably similar patterns during key economic phases, including crisis periods and growth cycles. This visual comparison highlights how global financial markets often move in tandem, reflecting shared responses to worldwide economic events.

\begin{figure}[h!]
    \centering
    \includegraphics[width=0.9\linewidth]{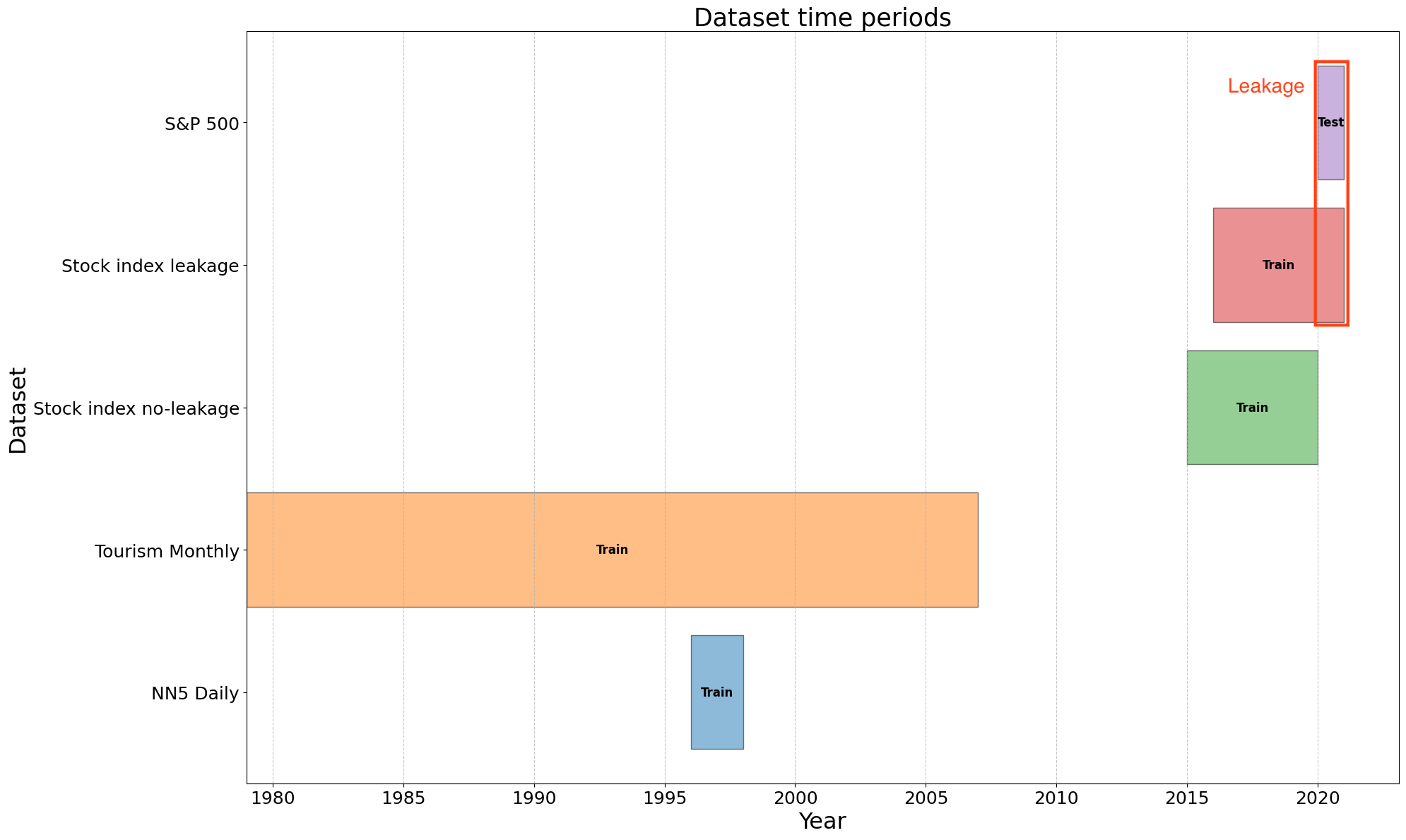}
    \caption{Comparison of time periods per dataset}
    \label{fig:time_datasets}
\end{figure}

Figure \ref{fig:time_datasets} illustrates the relationship between training and test data that overlap temporally. It is important to note that although the training and test periods overlap in time, they do not share the same data points. The overlap is purely temporal, meaning that both datasets cover a small portion of the same time period but contain different observations or series, as shown in Table \ref{tab:leakage_summary}.

\begin{table}[h!]
\centering
\begin{tabular}{@{}lcccc@{}}
\toprule
\multirow{2}{*}{Category} & \multicolumn{2}{c}{No Leakage} & \multicolumn{2}{c}{Indirect Leakage} \\ 
\cmidrule(lr){2-3} \cmidrule(lr){4-5}
 & Train & Test & Train & Test \\ 
\midrule
\# Time Series         & 484   & 1   & 484   & 1   \\ 
\# Data Points         & 205,854 & 230 & 205,859 & 230 \\ 
\# Indirect Leakage Data Points    & -     & -   & 1,600   & -   \\
\bottomrule
\end{tabular}
\caption{Summary of time series and data points with and without leakage}
\label{tab:leakage_summary}
\end{table}
\begin{table}[h!]
\centering
\caption{Per-seed S\&P 500 MAE breakdown for indirect leak from other countries stock indices vs no-leak experiments. The percentage difference shows how much better the model with leakage data performs than the non-leakage model. For each seed, the better MAE is displayed in bold.}
\label{tab:exp_single_seeds}
\begin{tabular}{ccccccc}
\toprule
\textbf{Seed} & \multicolumn{3}{c}{\textbf{MAE Before Split}} & \multicolumn{3}{c}{\textbf{MAE After Split}} \\
\cmidrule(lr){2-4} \cmidrule(lr){5-7}
              & \textbf{No Leak} & \textbf{Indirect Leak} & \textbf{\% Diff} & \textbf{No Leak} & \textbf{Indirect Leak} & \textbf{\% Diff} \\
\midrule
410 & 230.53 & \textbf{227.72} & 1.22\% & \textbf{157.85} & 199.33 & -26.28\% \\
1425 & 219.68 & \textbf{178.09} & 18.93\% & 120.77 & \textbf{82.53} & 31.67\% \\
1680 & 193.52 & \textbf{187.19} & 3.27\% & \textbf{80.74} & 81.22 & -0.60\% \\
1825 & 247.76 & \textbf{156.12} & 36.99\% & 110.47 & \textbf{70.36} & 36.31\% \\
2287 & 231.19 & \textbf{206.28} & 10.78\% & 102.42 & \textbf{100.54} & 1.84\% \\
3658 & 219.23 & \textbf{156.91} & 28.43\% & 117.25 & \textbf{84.62} & 27.84\% \\
4013 & 224.75 & \textbf{193.16} & 14.06\% & 114.19 & \textbf{97.98} & 14.20\% \\
4507 & \textbf{206.45} & 213.70 & -3.51\% & 93.10 & \textbf{84.01} & 9.77\% \\
8936 & \textbf{189.40} & 199.38 & -5.27\% & \textbf{103.59} & 153.67 & -48.35\% \\
9675 & 206.30 & \textbf{182.99} & 11.30\% & \textbf{82.35} & 132.44 & -60.82\% \\
\midrule
\textbf{Mean} & 216.88 & \textbf{190.15} & 11.62\% & \textbf{108.27} & 108.67 & -1.44\% \\
\textbf{Std Dev} & 17.15 & 21.88 & -- & 21.10 & 38.83 & -- \\
\bottomrule
\end{tabular}
\end{table}

\newpage
\subsection{Data Preprocessing and Model Tuning}
\label{sec:tuning}

Min-max scaling was applied separately to each dataset to account for differences in scale across datasets. Each scaler was fitted on the respective training data and subsequently used to transform the remaining data. For the S\&P 500 dataset, the scaler was fitted explicitly on data from 2015 to the end of 2019 and then applied to the data from 2020.

The Time-Series Transformer was implemented via the tsai library\footnote{\href{https://timeseriesai.github.io/tsai}{https://timeseriesai.github.io/tsai}}.
Hyperparameter tuning was performed via Optuna\footnote{\href{https://github.com/optuna/optuna}{https://github.com/optuna/optuna}} for 50 trials to minimize the MAE on the validation set.
The training set consisted of the first 80\% of NN5 Daily and Tourism Monthly datasets as well as January 1st, 2015 until June 30th, 2019 for no-leak and January 1st 2016 until June 30th, 2020 of the stock index data for leak.
The validation set consisted of the remaining 20\% of NN5 Daily and Tourism Monthly datasets, and the stock index data from July 1st, 2019 until December 31st, 2019 for no-leak and July 1st, 2020 until December 31st 2020 for leak.

\begin{table}[h!]
    \centering
    \resizebox{\textwidth}{0.1\textheight}{
    \begin{tabular}{llllll}
    \toprule
    \textbf{Parameter} & \textbf{Lower} & \textbf{Upper} & \textbf{Stepsize} & \textbf{Choice} \\
    \midrule
    Learning rate & log(1e-05) & log(1e-01) & - & - \\
    Batch size & - & - & - & 16, 32, 64, 128 \\
    Epochs & 2 & 20 & 1 & - \\
    Dropout & 0.0 & 0.7 & - & - \\
    FC Dropout & 0.0 & 0.7 & - & - \\
    Number of layers & 2 & 8 & 1 & - \\
    Number of heads & - & - & - & 4, 8, 16, 32 \\
    Model dimension & - & - & - & 64, 128, 256, 512, 1024 \\
    Dimension of feedforward network model & - & - & - & 32, 64, 128, 256, 512 \\
    \bottomrule
    \end{tabular}}
\caption{Optuna search space for tuning}
\label{tab:search_space_first_it}
\end{table}

\begin{table}[h!]
    \centering
    \resizebox{\textwidth}{!}{%
    \begin{tabular}{r r r r r r r r r r}
    \toprule
    \textbf{Seed} & \textbf{Learning rate} & \textbf{Batch size} & \textbf{Epochs} & \textbf{Dropout} & \textbf{FC Dropout} & \textbf{Number of layers} & \textbf{Number of heads} & \textbf{$d_\text{model}$} & \textbf{$d_\text{ff}$} \\
    \midrule
    410  & 9.687e-04 &  64 & 11 & 0.272 & 0.572 & 6 & 16 & 128 &  64 \\
    1425 & 5.889e-04 &  64 & 19 & 0.201 & 0.593 & 5 & 32 & 256 & 256 \\
    1680 & 5.773e-04 &  64 & 18 & 0.162 & 0.251 & 7 & 16 & 128 &  32 \\
    1825 & 4.392e-03 & 128 & 17 & 0.009 & 0.439 & 6 &  4 &  64 & 512 \\
    2287 & 4.828e-04 &  32 & 17 & 0.447 & 0.644 & 2 & 32 & 256 &  64 \\
    3658 & 3.502e-04 &  64 &  6 & 0.001 & 0.211 & 2 & 16 & 128 & 128 \\
    4013 & 5.348e-03 & 128 & 11 & 0.096 & 0.439 & 3 & 16 &  64 & 128 \\
    4507 & 3.656e-04 &  64 & 14 & 0.007 & 0.369 & 3 &  8 & 128 &  32 \\
    8936 & 5.428e-04 & 128 &  9 & 0.003 & 0.595 & 8 & 32 &  64 & 512 \\
    9675 & 3.102e-04 &  64 & 14 & 0.006 & 0.504 & 2 &  4 &  64 & 256 \\
    \bottomrule
    \end{tabular}
    }
\caption{Best hyperparameters per seed for no-leak models}
\label{table:best_params_no_leak}
\end{table}

\begin{table}[h!]
    \centering
    \resizebox{\textwidth}{!}{%
    \begin{tabular}{r r r r r r r r r r}
    \toprule
    \textbf{Seed} & \textbf{Learning rate} & \textbf{Batch size} & \textbf{Epochs} & \textbf{Dropout} & \textbf{FC Dropout} & \textbf{Number of layers} & \textbf{Number of heads} & \textbf{$d_\text{model}$} & \textbf{$d_\text{ff}$} \\
    \midrule
    410  & 1.851e-04 &  32 & 18 & 0.334 & 0.095 & 7 & 16 &  32 & 256 \\
    1425 & 1.156e-02 &  64 &  8 & 0.067 & 0.691 & 2 & 16 &  32 & 256 \\
    1680 & 3.105e-04 & 128 & 16 & 0.032 & 0.267 & 3 &  4 & 128 & 256 \\
    1825 & 8.714e-04 & 128 & 19 & 0.093 & 0.696 & 6 & 32 &  32 & 256 \\
    2287 & 2.703e-03 & 128 & 19 & 0.183 & 0.435 & 2 &  8 &  64 & 128 \\
    3658 & 1.645e-03 & 128 & 19 & 0.064 & 0.293 & 5 & 16 &  64 &  32 \\
    4013 & 2.590e-03 & 128 & 14 & 0.090 & 0.547 & 7 & 32 &  64 & 512 \\
    4507 & 4.160e-04 &  64 & 20 & 0.395 & 0.006 & 2 &  8 &  64 & 256 \\
    8936 & 1.328e-04 &  64 &  5 & 0.011 & 0.501 & 8 & 16 & 128 & 512 \\
    9675 & 3.373e-04 & 128 & 13 & 0.010 & 0.428 & 3 &  4 &  64 & 256 \\
    \bottomrule
    \end{tabular}
    }
\caption{Best hyperparameters per seed for leak models}
\label{table:best_params_leak}
\end{table}

\newpage
